%% file: main.tex
\documentclass[conference,a4paper]{IEEEtran}
\IEEEoverridecommandlockouts

\usepackage[hidelinks]{hyperref}
\usepackage[cmex10]{amsmath}
\usepackage{amssymb,amsfonts}
\usepackage{dblfloatfix}

\usepackage[ruled,vlined]{algorithm2e}
\usepackage{graphicx}

\usepackage{booktabs}
\usepackage{siunitx}
\usepackage[numbers,compress]{natbib}
\usepackage{texnames}
\usepackage{bm,bbm}
\usepackage{orcidlink}
\usepackage{subcaption}
\usepackage{mwe}
\usepackage{hyperref}
\usepackage{amsmath}
\usepackage{cleveref}

\DeclareMathOperator*{\argmin}{arg\,min}

\hypersetup{
    colorlinks=true,
    linkcolor=blue,
    filecolor=magenta,      
    urlcolor=cyan,
    pdftitle={Overleaf Example},
    pdfpagemode=FullScreen,
    }

\begin{document}

\title{Efficient Geometry-Controlled High-Resolution Satellite Image Synthesis
}

\author{\IEEEauthorblockN{Vlad Vasilescu\orcidlink{0000-0003-3743-4235}}
	\IEEEauthorblockA{\textit{Univ. POLITEHNICA Bucharest}\\
	\href{https://www.sigmalab.ro}{SIGMA Lab}, \href{https://campus.upb.ro}{CAMPUS Institute}\\
	\url{vlad.vasilescu2111@upb.ro}}
	\and
	\IEEEauthorblockN{Daniela Faur\orcidlink{0000-0001-5208-5753}}
	\IEEEauthorblockA{\textit{Univ. POLITEHNICA Bucharest}\\
	\href{https://geosense.ro}{GEOSENSE}, \href{https://campus.upb.ro}{CAMPUS Institute}\\
	\url{daniela.faur@upb.ro}}
	\and
	\IEEEauthorblockN{Teodor Costăchioiu\orcidlink{0009-0005-3564-3238}}
	\IEEEauthorblockA{\textit{Univ. POLITEHNICA Bucharest}\\
	\href{https://geosense.ro}{GEOSENSE}, \href{https://campus.upb.ro}{CAMPUS Institute}\\
	\url{teodor.costachioiu@upb.ro}}
}

\maketitle
\begin{abstract}
    High‑resolution satellite images are often scarce and costly, especially for remote areas or infrequent events. This shortage hampers the development and testing of machine‑learning models for land‑cover classification, change detection, and disaster monitoring. In this paper, we tackle the problem of geometry-controlled high-resolution satellite image synthesis by adding control over existing pre-trained diffusion models. We propose a simple yet efficient method for controlling the synthesis process by leveraging only skip connection features using windowed cross-attention modules. Several previously established control techniques are compared, indicating that our method achieves comparable performance while leading to a better alignment with the geometry control map. We also discuss the limitations in current evaluation approaches, amplifying the necessity of a consistent alignment assessment. Our model and inference pipeline can be found at \href{https://github.com/Vladimirescu/EfficientGeometrySatelliteSynthesis.git}{Vladimirescu/EfficientGeometrySatelliteSynthesis}.
\end{abstract}


\input{sections/introduction}
\input{sections/problemform}
\input{sections/experiments}
\input{sections/conclusions}

\small
\bibliographystyle{IEEEtranN}
\bibliography{references}

\end{document}

%% file: sections/introduction.tex
\section{Introduction}

Satellite image synthesis offers a realistic solution to augment datasets \cite{adedeji2022image}, simulate various scenarios \cite{rui2021disastergan, lagap2025predicting}, and provide missing sensor modalities, enabling research that would otherwise be infeasible \cite{panchal2021reconstruction}. With the current surge of powerful generative models \cite{labs2025flux, rombach2022high, podell2023sdxl}, able to seemingly control and edit visual content through various conditions, the endeavor of high resolution remote sensing data generation has become more feasible. Applications such as multispectral image synthesis \cite{alibani2024multispectral, abady2020gan}, optical image generation \cite{sastry2024geosynth, shah2021satgan} and change detection \cite{zheng2024changen2} comprise a small subset of remote sensing tasks which have benefited from the advent of generative models. 

Although pre-trained generative models have shown remarkable capabilities across a wide range of tasks \cite{rombach2022high, labs2025flux}, this generality comes at the cost of precision and task-specific performance. To address this, a multitude of works has focused on incorporating control mechanisms \cite{liu2024smartcontrol, zhao2023uni}, such as sketches \cite{voynov2023sketch}, depth or segmentation maps \cite{zhang2023adding}, specializing the model to meet particular constraints. 

In this work, we investigate the problem of controlling pre-trained text-to-image (TTI) diffusion models to generate geometry-aware high-resolution satellite images. For the geometry, we consider tiles from Open Street Map (OSM), following the protocol in \cite{sastry2024geosynth}. In \Cref{sec:probform} we discuss the building blocks of such control mechanisms within related works, and propose our approach to an efficient control scheme. \Cref{sec:exps} offers an extensive evaluation of several techniques, showcasing the performance of our method and discussing the current limitations in assessing the alignment with guiding information. Finally, \Cref{sec:concl} concludes our study and highlights improvement directions.

%% file: sections/problemform.tex
\section{Problem Formulation}\label{sec:probform}

\subsection{Diffusion Models}

DDPMs \cite{ho2020denoising} are a class of generative models based on two processes: a \textit{forward diffusion process} $q(x_t|x_{t-1})$, which gradually perturbs clean images $x_0$, and a \textit{backward reverse process}, where an approximate reverse procedure $p(x_{t-1} | x_t)$ is searched for. For a fixed number of diffusion steps $T\in \mathbb{N}_+$, noisy observations in the forward process are created as $x_t = \sqrt{\bar{\alpha}_t} x_0 + \sqrt{1 - \bar{\alpha}_t} \eta_t$, $\eta_t \sim \mathcal{N}(\mathbf{0}, \mathbf{I})$, where $\bar{\alpha}_t = \prod_{i=1}^t (1 - \beta_i)$ and $\{\beta_t\}_{t=1}^T$ are the noise schedule variances. To approximate the backward process, a neural network denoiser $\epsilon_{\theta}$ parametrized by weights $\theta$ is trained to approximate noise $\widehat{\eta}_t = \epsilon_{\theta}(x_t, t)$ using standard squared $\ell_2$ loss. In recent applications, the denoising process is additionally guided by extra control information, such as text \cite{brack2024ledits}, segmentation maps \cite{zhao2023uni}, or 3D structured data \cite{luo20213d}, resulting in the following general optimization problem:
\begin{align}\label{eq:problem}
    \argmin_\theta \mathbb{E}_{(x_0, \texttt{*args}), t\sim\mathcal{U}(1, T)} \Vert \eta_t - \epsilon_{\theta}(x_t, t, \texttt{*args}) \Vert_2,
\end{align}
where \texttt{*args} denotes a list of available control information associated to the clean data sample $x_0$.

Latent Diffusion Models (LDMs) \cite{rombach2022high} addressed the challenge of efficiently generating high-resolution data by moving the diffusion process in the latent space spanned by extensively-trained encoder-decoder structures $\mathcal{E} \circ \mathcal{D}$. In this case, noisy observations are computed as $z_t = \sqrt{\bar{\alpha}_t} \mathcal{E}(x_0) + \sqrt{1 - \bar{\alpha}_t} \eta_t$, and the final prediction is mapped through the decoder in the original space $\hat{x}_0 = \mathcal{D}(\hat{z}_0)$. This methodology remains at the base of some of the most powerful generative models, such as Stable Diffusion (SD) \cite{rombach2022high}.

\subsection{Related Work: Control Mechanisms}

\begin{figure*}[t]
    \centering
    \begin{subfigure}{0.21\textwidth}
        \includegraphics[width=\linewidth]{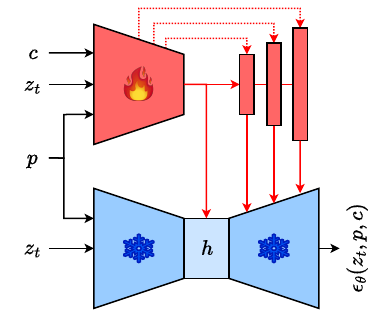}
        \caption{ControlNet \cite{zhang2023adding}}
        \label{fig:controlnet}
    \end{subfigure}
    \begin{subfigure}{0.23\textwidth}
        \includegraphics[width=\linewidth]{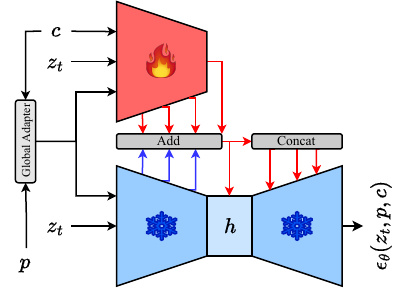}
        \caption{Uni-ControlNet \cite{zhao2023uni}}
        \label{fig:unicontrol}
    \end{subfigure}
    \begin{subfigure}{0.23\textwidth}
        \includegraphics[width=\linewidth]{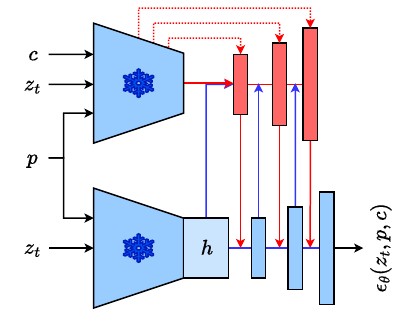}
        \caption{SmartControl \cite{liu2024smartcontrol}}
        \label{fig:smartcontrol}
    \end{subfigure}
    \begin{subfigure}{0.23\textwidth}
        \includegraphics[width=\linewidth]{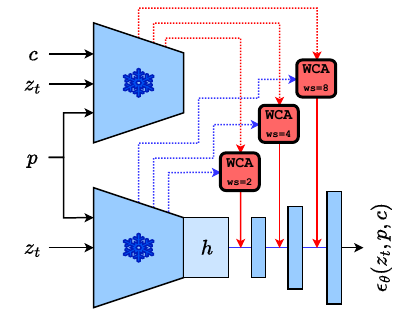}
        \caption{\textbf{Ours}}
        \label{fig:ours}
    \end{subfigure}
    \caption{Comparison between different control mechanisms and our proposed \textit{Windowed Cross-Attention} (WCA) approach. $p$ denotes the text prompt, $z_t$ the noisy sample, and $c$ the additional geometry control information.}
    \label{fig:archs}
\end{figure*}

Given the extensive training of currently open-source diffusion models, the need for efficient fine-tuning and the introduction of additional control signals remains an ongoing interest. Several works aimed to enhance control over diffusion-based generation at inference time without retraining \cite{voynov2023sketch, bar2023multidiffusion, bashkirova2023masksketch, mou2024t2i}, respecting the external guidance signals beyond text prompts. In the case of synthesizing satellite images, however, a significant distribution shift is required to account for specialized generation, deeming such test-time control methods insufficient. Fine-tuning methods \cite{hulora, ruiz2023dreambooth} represent a solution to account for significant data shifts, however they do not explicitly enforce strong geometry control, unless additional learnable modules are applied.

\textit{ControlNet} \cite{zhang2023adding} introduces additional control signals over a pre-trained diffusion model by creating a learnable copy of the model's encoder, forwarding the new condition tensors through it and using the up-sampled decoding features to control the decoding dynamics of the based diffusion model (see \Cref{fig:controlnet}). \textit{Zero convolutions} allow gradients to pass freely towards the learnable control modules, while keeping the performance equivalent to the base architecture in the initial training steps. \textit{Uni-ControlNet} \cite{zhao2023uni} uses a \textit{global adapter} to modify the original text prompt using the new control signal, and a \textit{local adapter} which combines the features from the original and a learnable copy of the encoder, to be further combined with based decoding features (see \Cref{fig:unicontrol}). \textit{SmartControl} \cite{liu2024smartcontrol} proposes a more lightweight control scheme, where the new control signals are forwarded through a freezed copy of the encoder, while the decoding features are combined through \textit{control scale predictors} which produce pixel-wise attention maps which guide the intertwining between original and control features (see \Cref{fig:smartcontrol}). GeoSynth \cite{sastry2024geosynth} is one of the first works to address controlled satellite image synthesis, using ControlNet at its core and a variety of guiding signals.

\subsection{Proposed Control Method}

With the aim of designing a lightweight and efficient control mechanism, we begin with adopting the freezed encoder copy strategy proposed by SmartControl \cite{liu2024smartcontrol}. In order to ensure alignment between control and base features, we adopt the usage of \textit{only} corresponding skip-connection information to control the entangling between control and base networks. We further denote the set of skip connection features computed by the base model as $\{s_i\}_{i=1}^N$, and by the control model as $\{s^{\rm ctr}_i\}_{i=1}^N$, where $N\in \mathbb{N}_+$ is the number of encoding layers. Given base decoding features $d_j$ at decoding block $j$, the control information is added as follows:
\begin{align}
    d_j \leftarrow [d_j \, ; \, s_{N-j} + \alpha_{N-j} \odot s_{N-j}^{\rm ctr}] 
\end{align}
where $\{\alpha_i\}_{i=1}^N$ are pixel-wise mixing maps with values within $[0, 1]$, and $[\cdot \, ; \, \cdot]$ denotes channel-wise concatenation. The base model can be recovered by setting all $\alpha_i = \mathbf{0}$. To compute $\alpha_i$, we adopt \textit{Windowed Cross-Attention} (WCA) \cite{vaswani2021scaling} modules between skip-connection features $s_i$ and $s^{\rm ctr}_i$ (\Cref{fig:ours}):
\begin{align}
    \text{WCA}_i &: \begin{cases}
        Q_i &= t_{\rm ws}(w^q_i \ast s_i^{\rm ctr}) \\
        K_i &= t_{\rm ws}(w^k_i \ast s_i) \\
        V_i &= t_{\rm ws}(w^v_i \ast s_i) 
    \end{cases}\rightarrow
    \begin{cases}
        A_i &= \text{SM}(Q_i K_i^\top)\\
        O_i &= t_{\rm ws}^{-1}(A_i V_i)
    \end{cases} \nonumber \\
    \alpha_i &= \sigma(w^{1\times 1}_i \ast O_i)
\end{align}
where $\text{SM}(\cdot)$ represents the \textit{Softmax} operator, $\sigma(\cdot)$ the \textit{Sigmoid}, and $t_{\rm ws}$ the windowing reshape operator with window size $\rm ws \geq 2$. $w^{1\times 1}_i$ is a $1\times 1$ convolutional kernel mapping $O$ to a single-channel output. This choice is justified by the necessity of fine-grained control within each spatial window, avoiding interference from any global, possibly overpowering control features. This way, each generated spatial region should become more aligned with its corresponding region within the control signal. 

%% file: sections/experiments.tex
\section{Experiments}\label{sec:exps}

\subsection{Setup}

\textit{Data.} We apply the same protocol described in GeoSynth \cite{sastry2024geosynth}, using their dataset comprised of $512\times 512$ RGB satellite images $x_0$ with a ground sampling distance of $0.6m$. Each image has a corresponding textual description $p$ and an Open Street Map (OSM) \footnote{\url{www.openstreetmap.org}} tile $c$. As there are no official validation or test sets available, we report all our metrics on the first $3000$  samples in the dataset, for all considered models.

\textit{Control Models.} We tested GeoSynth \cite{sastry2024geosynth}, which is based on the ControlNet \cite{zhang2023adding} architecture. Next, we tested Uni-ControlNet \cite{zhao2023uni}, for which we adopted its version with local control only, since the global adapter is usually applied when there are multiple control signals to be accounted for. For SmartControl \cite{liu2024smartcontrol} we tested both the original version, and another with \textit{Convolutional Block Attention Modules} (CBAM) \cite{woo2018cbam} instead of its original control scale predictors. For our proposed WCA-controlled architecture we tried several window size configurations to discuss the optimal setting. Stable Diffusion \cite{rombach2022high} is used as the base / controlled model along each control method.

\textit{Training}. We train all the control methods from scratch, using similar training hyperparameters to ensure a fair comparison. Models are trained for $20$ epochs, with a learning rate of $10^{-5}$ and the based Stable Diffusion models completely freezed. The optimization is performed according to Equation \eqref{eq:problem}, choosing $\texttt{*args}=[p, c]$, i.e. the textual description and the OSM tile. 

\textit{Metrics.} To assess the quality of synthesized satellite images, we use FID and CLIP to measure distributional similarity and semantic guidance, respectively. We do not use SSIM as it doesn't account for relevant information in fully-generative scenarios. Additionally, we use CLIP-IQA \cite{wang2023exploring} to measure the quality assessment of generated images, with an antonym prompt strategy that leverages CLIP’s prior knowledge. Five characteristics for CLIP-IQA are considered: \textit{quality}, \textit{sharpness}, \textit{contrast}, \textit{naturalness} and \textit{realness}. Finally, we acknowledge that it is challenging to design a metric to measure the alignment between a synthesized image and the conditional OSM tile, leaving this as an open problem. 

\subsection{Results \& Discussion}

\input{tables/main}

\begin{figure}[t]
    \centering
    \includegraphics[width=0.8\linewidth]{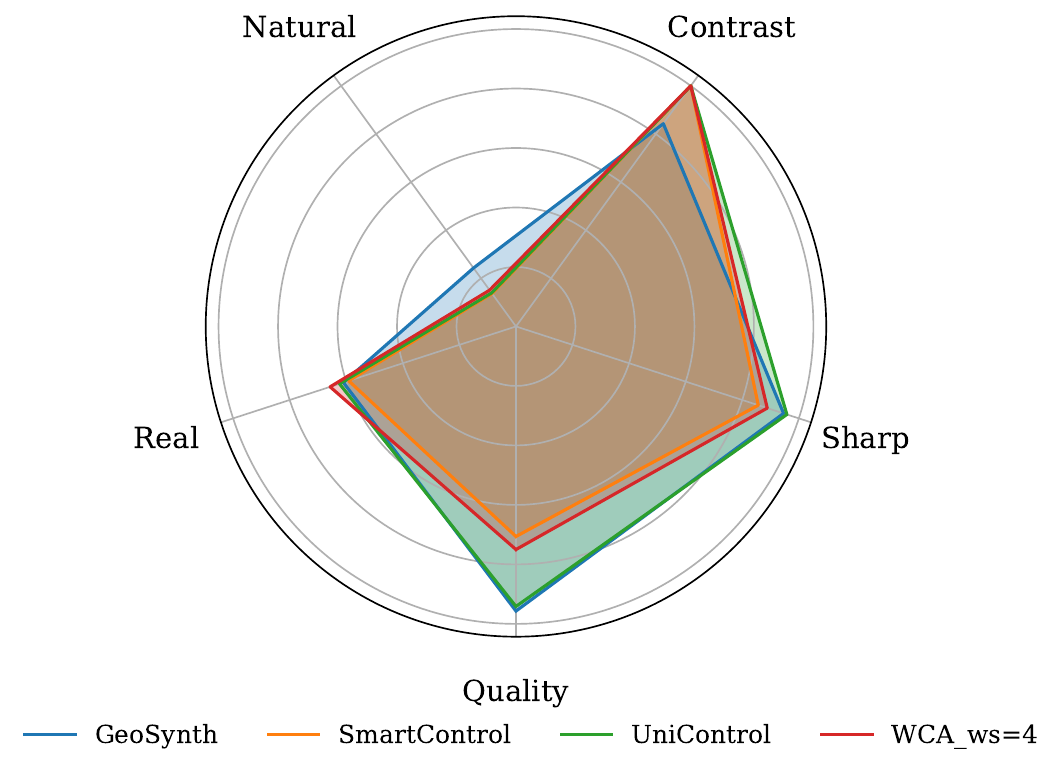}
    \caption{CLIP-IQA results for the 5 considered characteristics. Values are reported with respect to the \textbf{real data} metrics.}
    \label{fig:clip-iqa}
\end{figure}

\input{tables/runtime}

\input{Figures/generation_examples}

Table \ref{tab:main} showcases the synthesis results for all considered control models. Figure \ref{fig:gen-examples} also showcases several examples generated by each method.

First thing to note is that, in terms of semantic alignment (CLIP), all models result in roughly the same performance, indicating a limitation of pre-trained CLIP models to capture image-text consistencies in more specialized scenarios, e.g., high-resolution nadiral image captioning. This is particularly shown by the average CLIP on real satellite images, indicating its lack of comprehensiveness in these scenarios.

In terms of distributional similarity (FID \& LPIPS), Uni-ControlNet results in the worst performance, highlighting its low capability to capture realistic features within satellite imagery. Another plausible interpretation could be that this approach doesn't sufficiently account for the control OSM, resulting in generated visual content completely unaligned with the reference information. This is showcased through the examples from Figure \ref{fig:gen-examples}, with Uni-ControlNet having the lowest image-control alignment. On another note, our WCA-control model and SmartControl achieve the best FID and LPIPS, resulting in both natural synthesized features and visually good alignment with OSM.

In the case of our proposed method, it can be observed that lower window sizes result in better generation capabilities, indicating that higher attention windows result in undesired interference between control features and the base model. Additionally, we observe that having a fixed window size for all decoding blocks (e.g. $\rm ws=4$) vs. gradually increasing it with the decoder block resolution (e.g. $\rm ws=[2, 4, 8]$) does not lead to a significant difference in performance. This indicates that both a constant low or a resolution-adapted low window size can yield good alignment.

For image quality assessment (CLIP-IQA), Table \ref{tab:main} shows the average results across all 5 considered characteristics, for both generated and real data. In Figure \ref{fig:clip-iqa} we include a radar plot to independently assess each attribute, illustrating the metrics obtained on generated data with respect to the ones obtained on real satellite images. Uni-ControlNet achieves the highest average CLIP-IQA, mainly due to its high \textit{quality} and \textit{sharpness} scores, compared to the other control mechanisms. This is also observed in Figure \ref{fig:gen-examples}, with Uni-ControlNet images having  significantly more details. It falls short, however, in terms of \textit{naturalness}, indicating misalignment with natural satellite imagery features. On the other hand, GeoSynth / ControlNet has the highest level of \textit{naturalness}, while falling short in terms of \textit{contrast}. Our WCA-control model outperforms SmartControl on all $5$ attributes, resulting in a better alignment with real satellite imagery. Finally, it is worth noting that all methods fall short w.r.t naturalness and realness, suggesting limitations in capturing the realistic attributes of satellite data.

In terms of sampling efficiency, \Cref{tab:runtime} reports the average time, memory, and FLOPs for inference sampling. While GeoSynth results in the overall most efficient system, it lacks in generative performance. On the other hand, SmartControl obtains significantly better performance, at the cost of high computational cost and model size. Our proposed mechanism results in a good trade-off between inference efficiency and high-quality generation.

In the end, the lack of a robust alignment metric between an information-sparse control signal (OSM) and a highly detailed synthesized image remains an open problem, deepening as we continuously push toward highly specialized generative models. In terms of visual quality assessment (Figure \ref{fig:gen-examples}), SmartControl and our WCA proposed control model lead to significantly more aligned images, indicating that simple, well-constructed control techniques represent an optimal choice. Future research should focus on whether these alignment strategies can scale to even more complex, multi-modal control signals without sacrificing the creative flexibility inherent in diffusion-based models.

%% file: tables/main.tex
\begin{table}[t]
\centering
\resizebox{\linewidth}{!}{%
\begin{tabular}{lcccc}\toprule \midrule
               & \textbf{FID}$\downarrow$ & \textbf{LPIPS}$\downarrow$ & \textbf{CLIP}$\uparrow$ & \textbf{CLIP-IQA}$\uparrow$ \\ \midrule
\textbf{real data} & $0$ & $0$ & $0.19$ & $0.745$ \\ \midrule
ControlNet     & $54.64$ & $0.611$ & $0.19$ & \underline{$0.681$} \\
Uni-ControlNet & $66.31$ & $0.621$ & $0.19$ & $\mathbf{0.696}$ \\
SmartControl   & $\mathbf{47.37}$ & $0.615$ & $0.19$ & $0.666$ \\
SmartControl$_{\rm CBAM}$   & $48.13$ & \underline{$0.605$} & $0.19$ & $0.653$ \\ \midrule 
WCA$_{\rm ws=8}$  & $52.27$ & $0.611$ & $0.19$ & $0.673$ \\ 
WCA$_{\rm ws=4}$  & \underline{$47.83$} & $0.615$ & \underline{$0.20$} & $0.674$ \\
WCA$_{\rm ws=[4, 8, 16]}$  & $50.33$ & $0.612$ & \underline{$0.20$} & $0.669$ \\ 
WCA$_{\rm ws=[2, 4, 8]}$  & $49.94$ & $\mathbf{0.604}$ & $\mathbf{0.21}$ & $0.656$ \\
\midrule \bottomrule
\end{tabular}
}
\caption{Performance comparison between different control mechanism for satellite image synthesis.}
\label{tab:main}
\end{table}

%% file: tables/runtime.tex
\begin{table}[t]
\resizebox{\linewidth}{!}{%
\begin{tabular}{llccc}\toprule
 & size (M) & Time/batch $[\texttt{ms}]$ & Max. VRAM $[\texttt{GB}]$ & TFLOPs \\ \midrule
ControlNet/GeoSynth & $865$ & $\phantom{0}923 \pm 1.1$ & $8.5$ & $43.52$ \\
UniControlNet & $859$ & $1121 \pm 0.1$ & $8.9$ & $45.28$ \\
SmartControl & $1581$ & $1384 \pm 5.4$ & $11.5$ & $66.27$ \\
WCA$_{\texttt{ws}=4}$ & $893$ & $1146 \pm 3.6$ & $8.9$ & $44.33$ \\ \bottomrule
\end{tabular}
}
\caption{Inference runtime statistics for a batch size of $4$ and \texttt{DDIM} sampler with $20$ steps.}
\label{tab:runtime}
\end{table}

%% file: Figures/generation_examples.tex
\begin{figure*}[t]
    \centering
    \begin{subfigure}[t]{0.19\textwidth}
        \centering
        \caption*{OSM Control}
        \includegraphics[width=\linewidth]{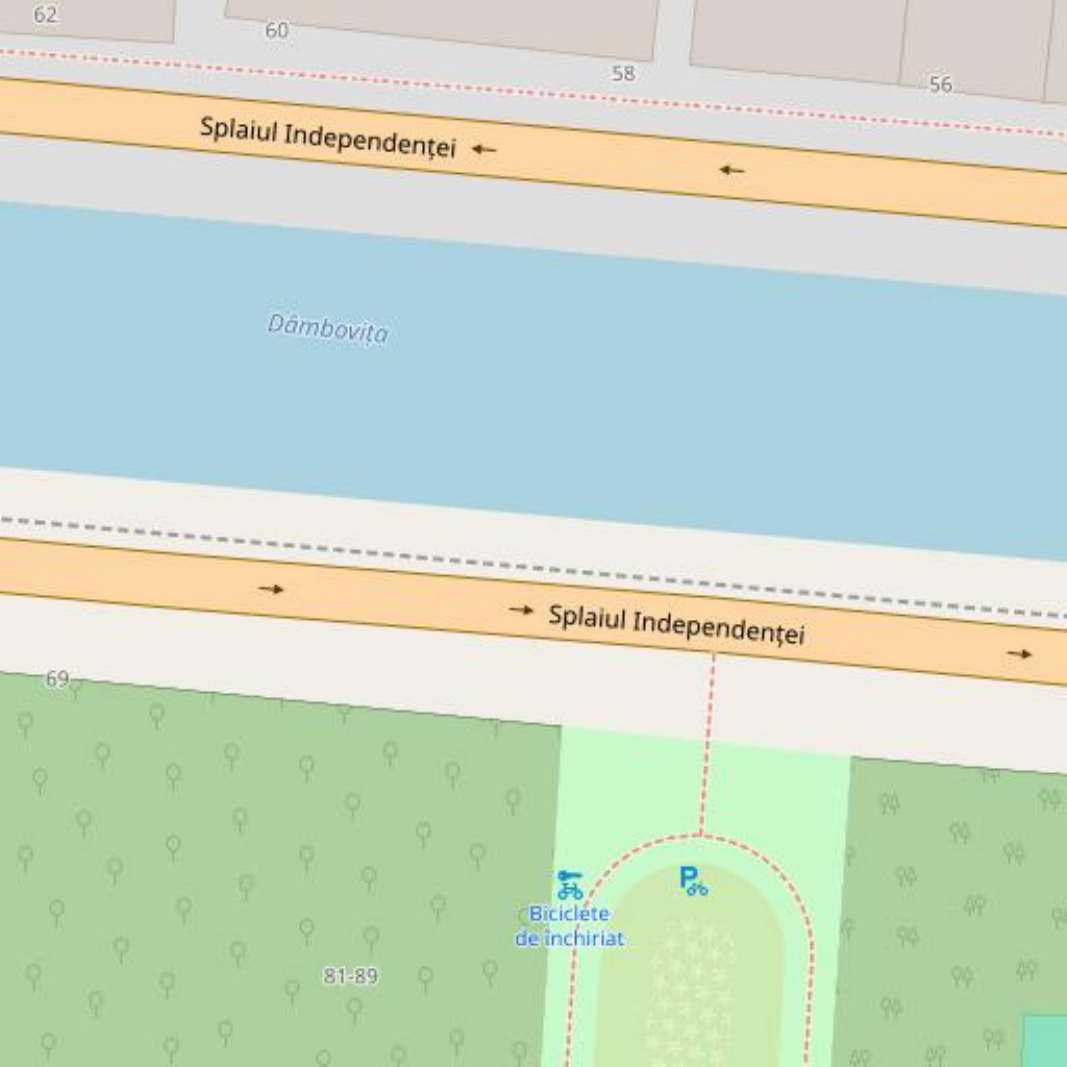}
    \end{subfigure}
    \hfill
    \begin{subfigure}[t]{0.19\textwidth}
        \centering
        \caption*{GeoSynth (ControlNet)}
        \includegraphics[width=\linewidth]{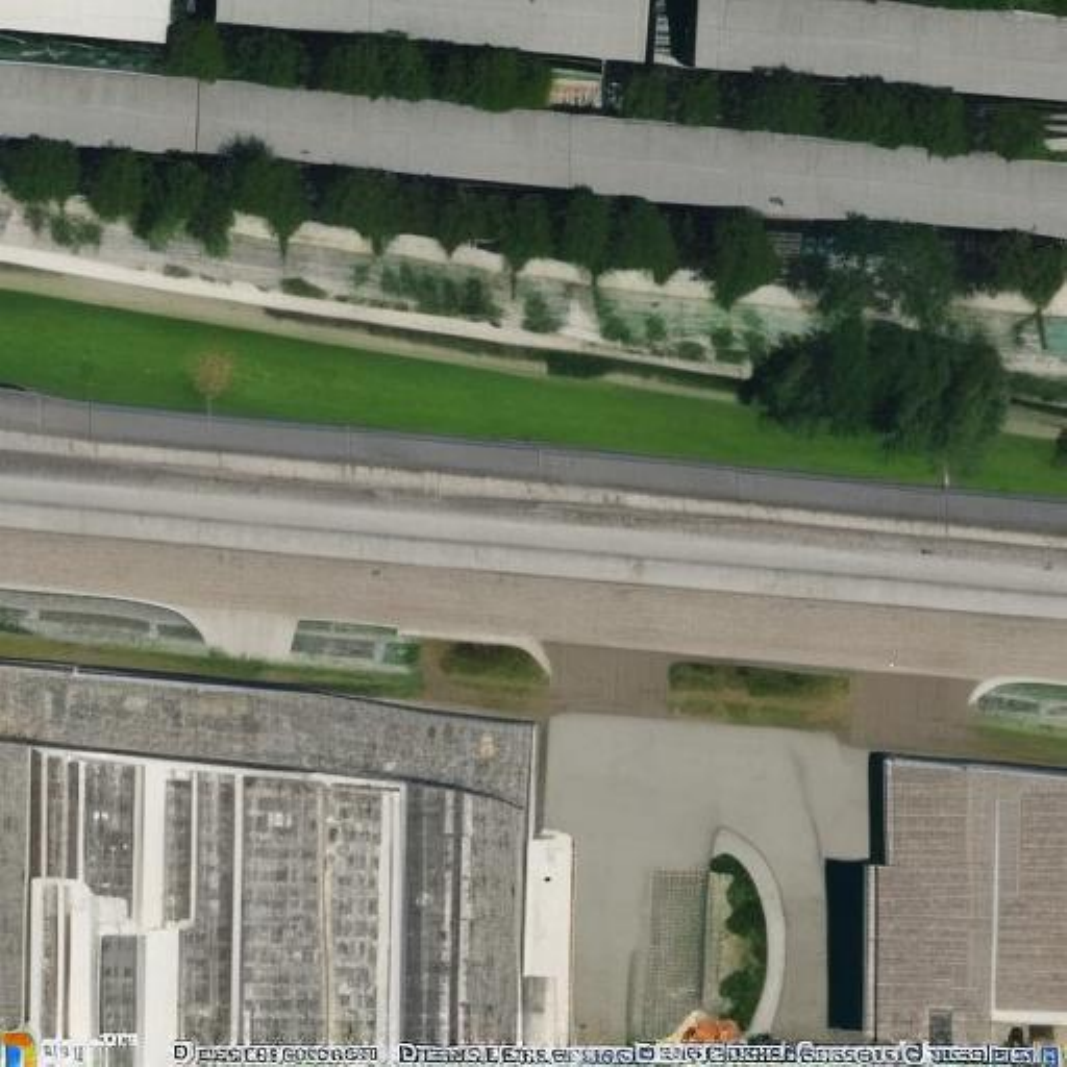}
    \end{subfigure}
    \hfill
    \begin{subfigure}[t]{0.19\textwidth}
        \centering
        \caption*{Uni-ControlNet}
        \includegraphics[width=\linewidth]{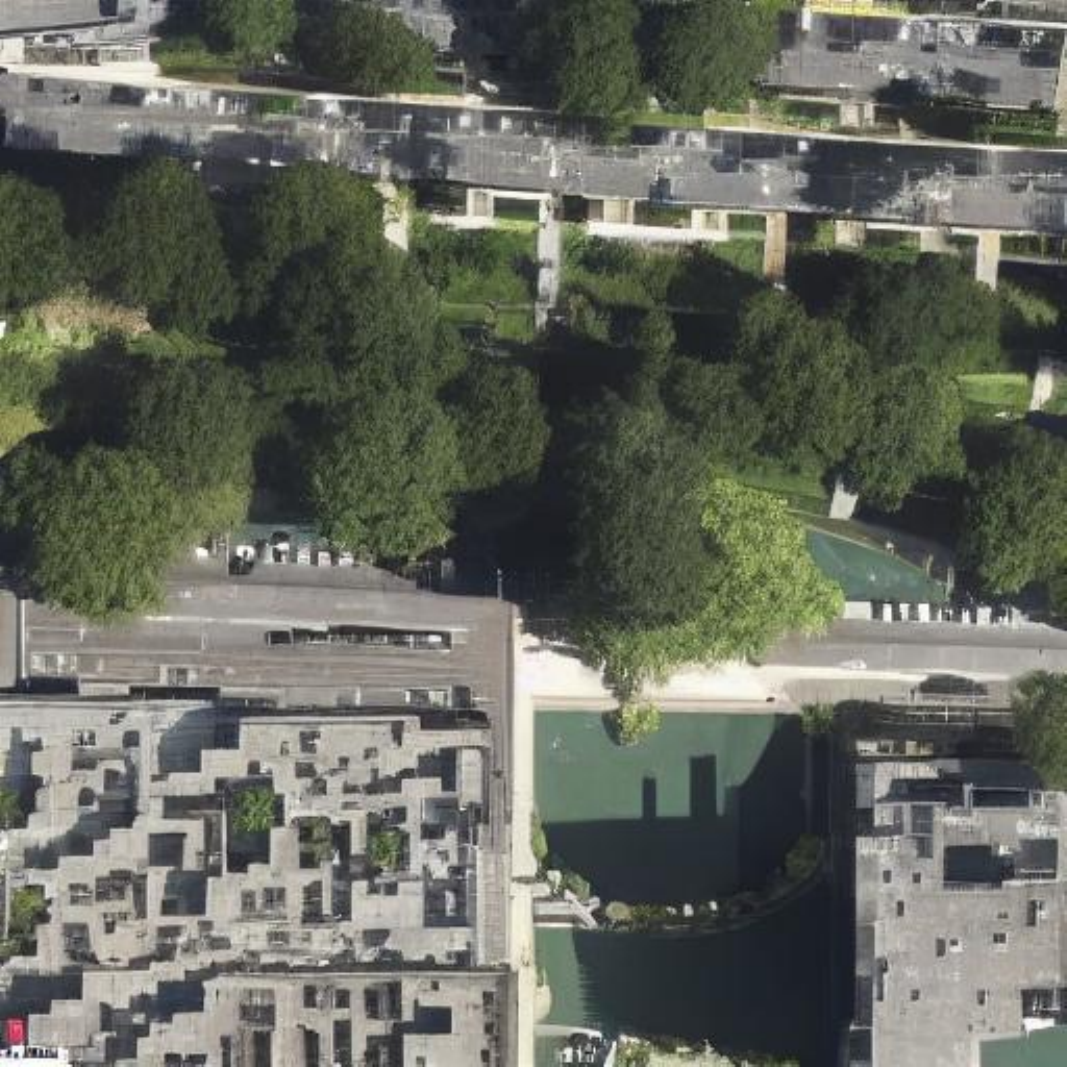}
    \end{subfigure}
    \hfill
    \begin{subfigure}[t]{0.19\textwidth}
        \centering
        \caption*{SmartControl}
        \includegraphics[width=\linewidth]{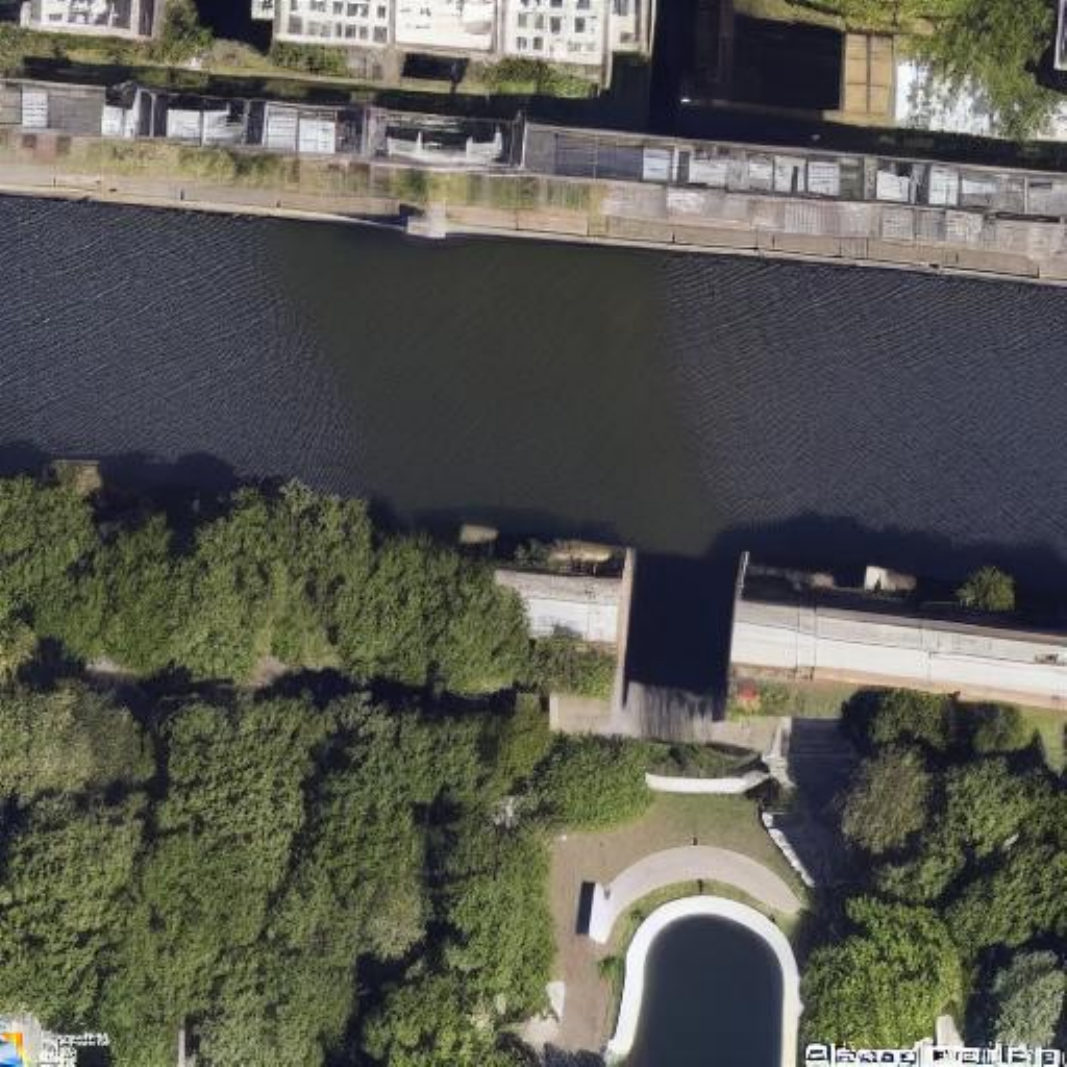}
    \end{subfigure}
    \hfill
    \begin{subfigure}[t]{0.19\textwidth}
        \centering
        \caption*{WCA$_{\rm ws=4}$ (\textbf{Ours})}
        \includegraphics[width=\linewidth]{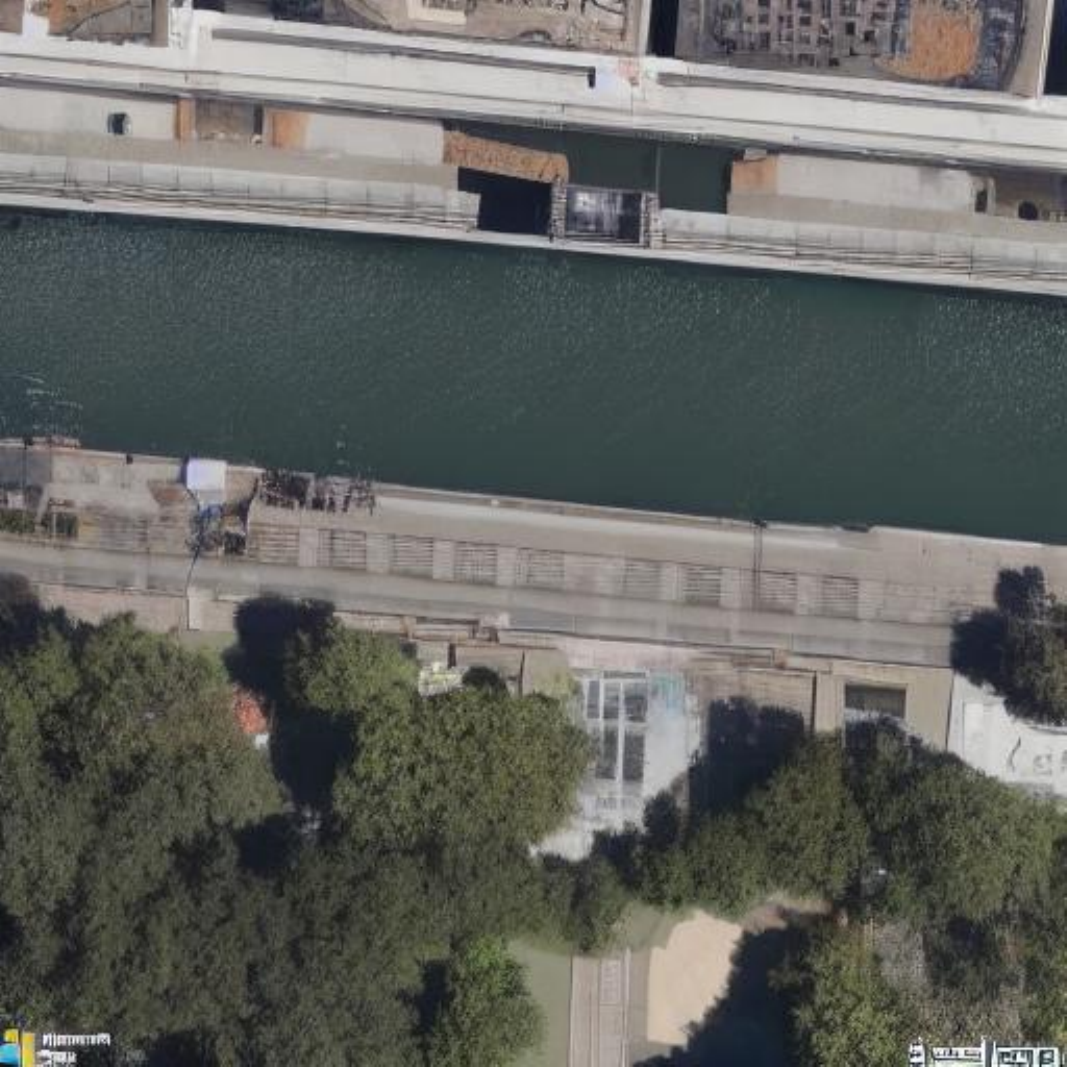}
    \end{subfigure}
    \par\medskip
    {\footnotesize\textit{"An urban area with a clear-water river, a park and blocks."}}
    \par\medskip
    \begin{subfigure}[t]{0.19\textwidth}
        \centering
        \includegraphics[width=\linewidth]{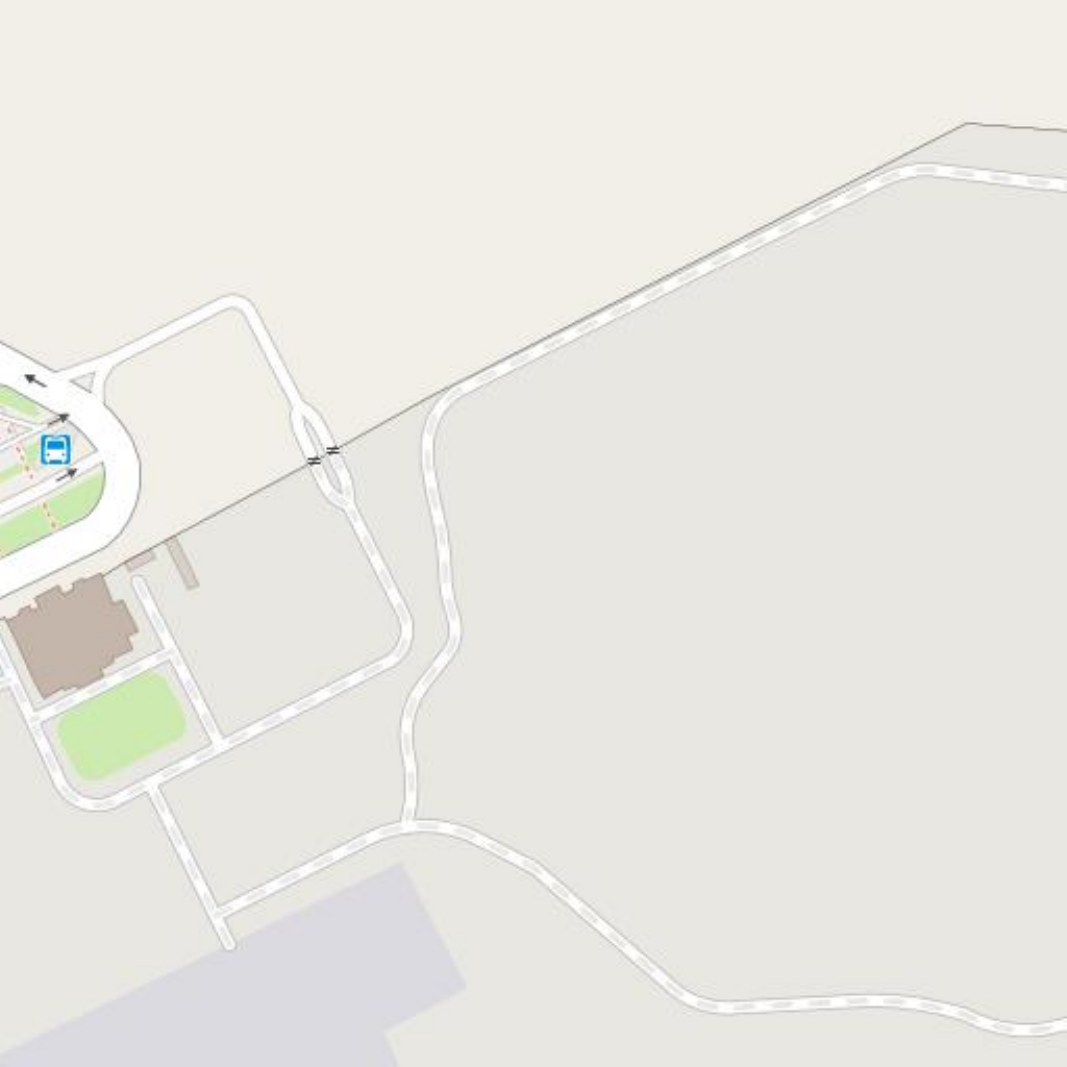}
    \end{subfigure}
    \hfill
    \begin{subfigure}[t]{0.19\textwidth}
        \centering
        \includegraphics[width=\linewidth]{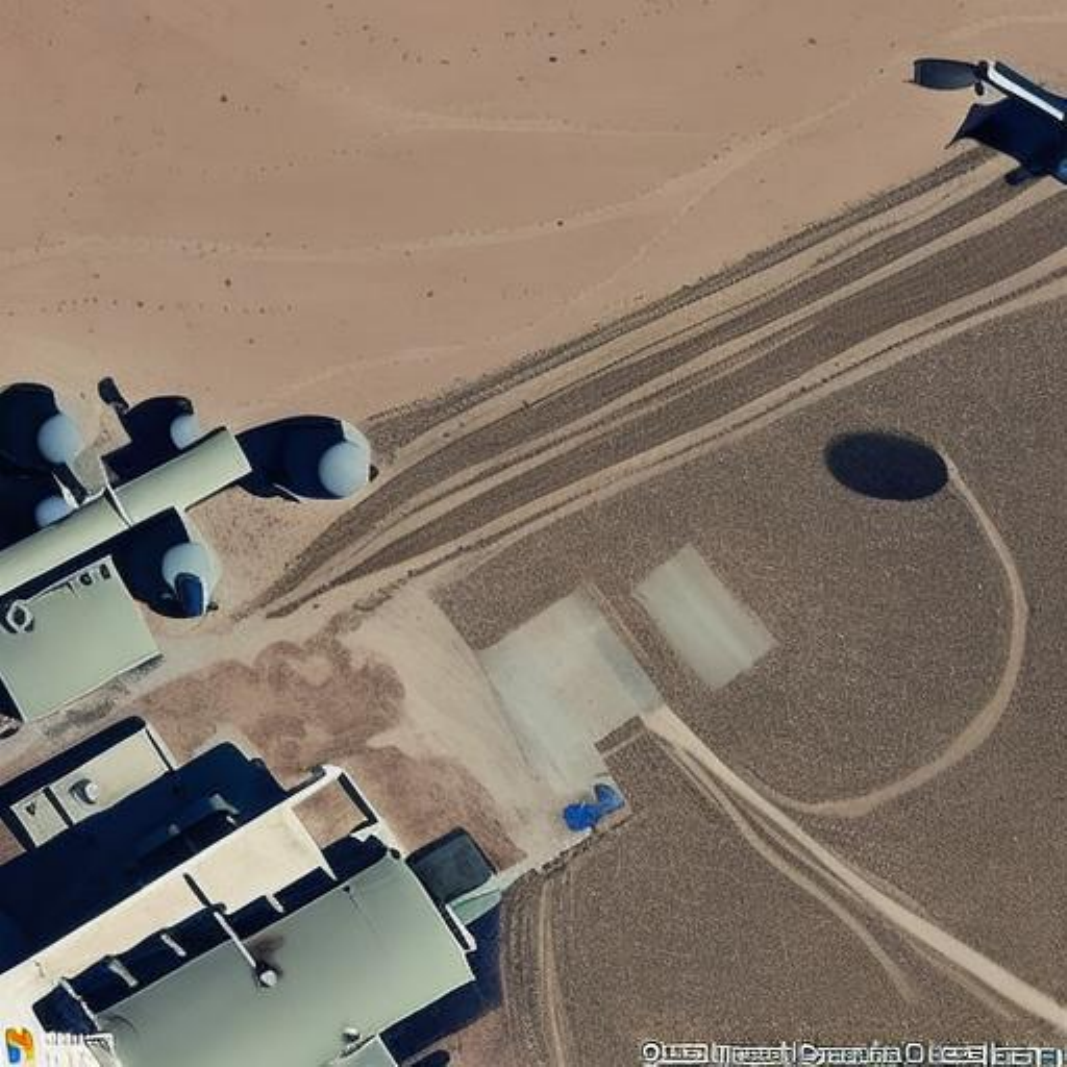}
    \end{subfigure}
    \hfill
    \begin{subfigure}[t]{0.19\textwidth}
        \centering
        \includegraphics[width=\linewidth]{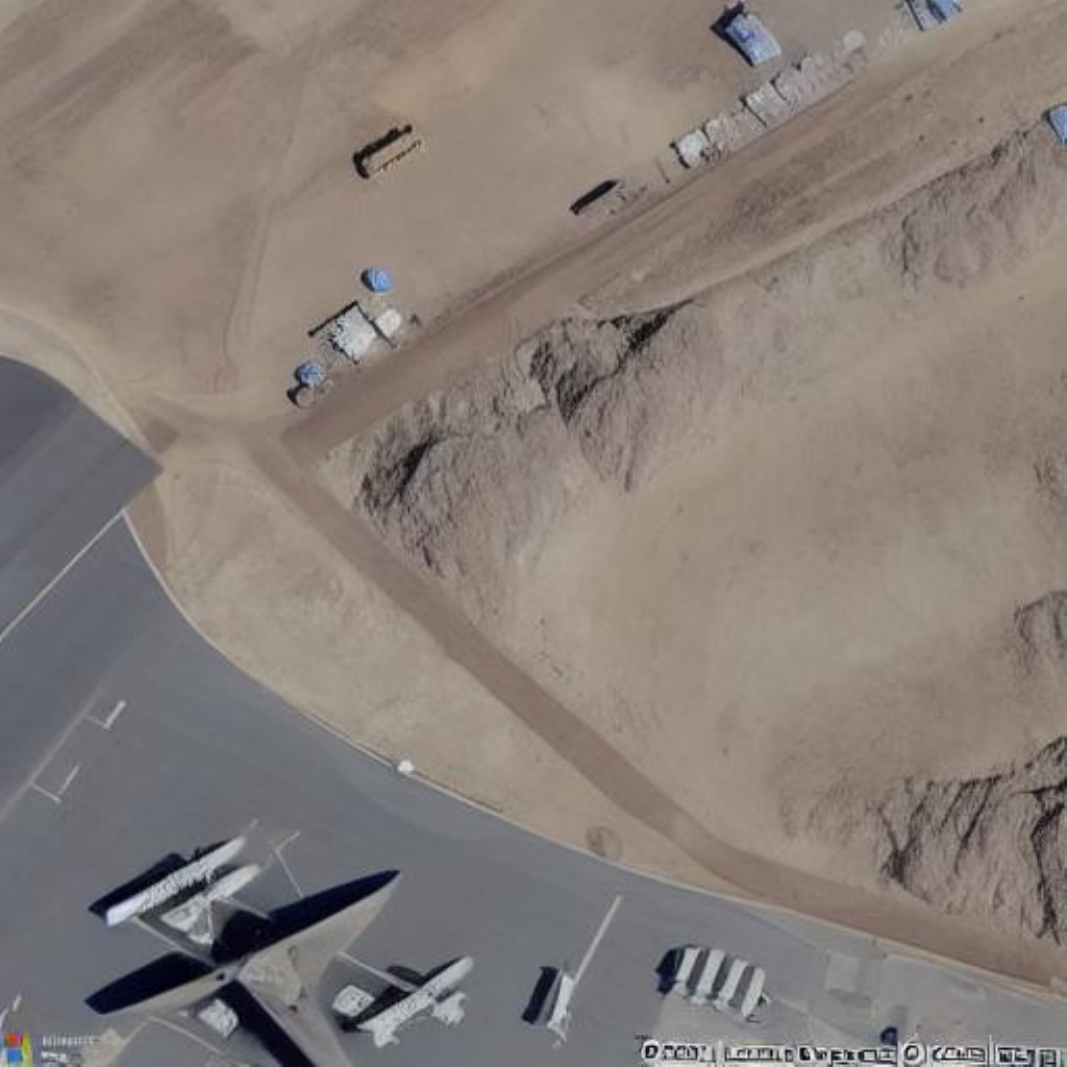}
    \end{subfigure}
    \hfill
    \begin{subfigure}[t]{0.19\textwidth}
        \centering
        \includegraphics[width=\linewidth]{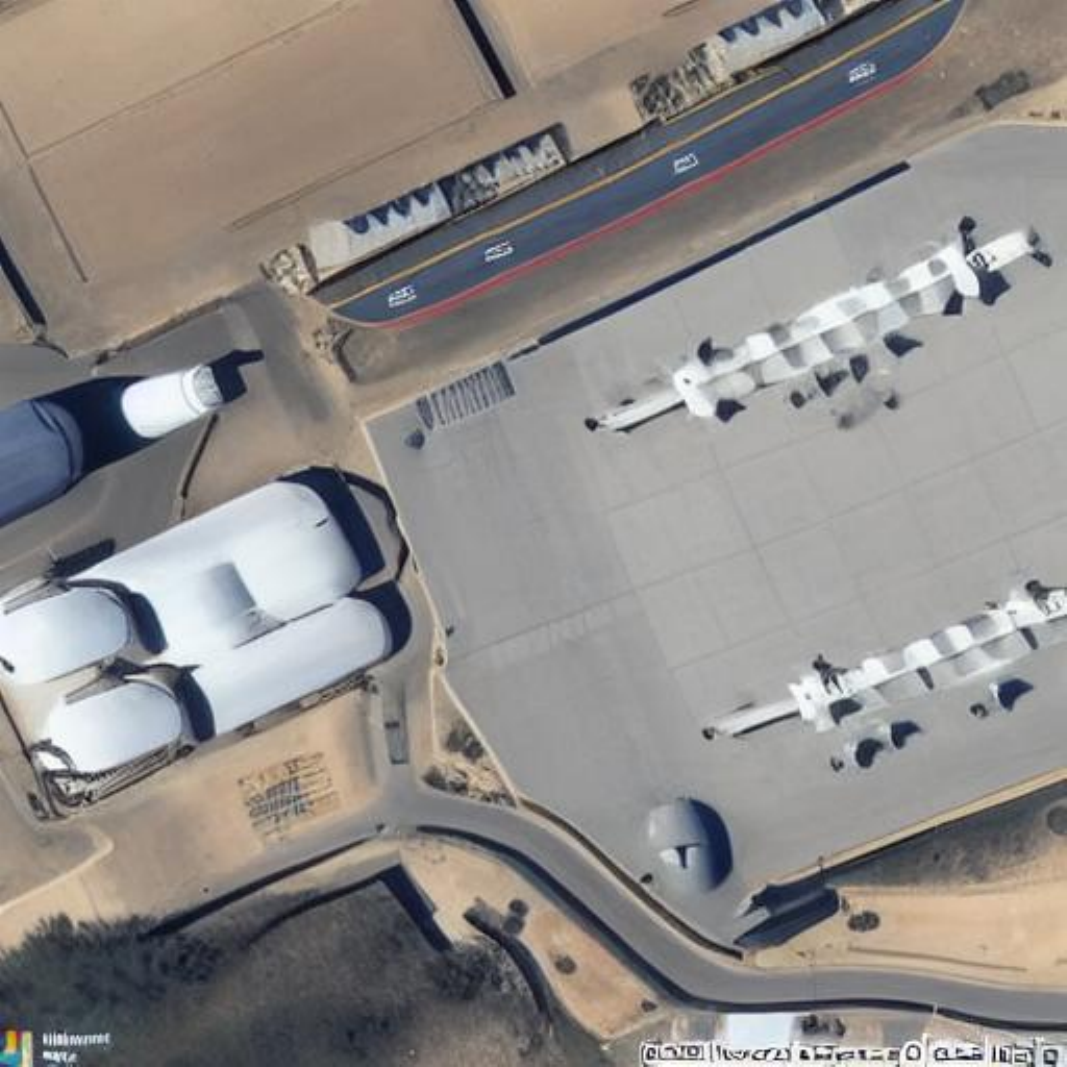}
    \end{subfigure}
    \hfill
    \begin{subfigure}[t]{0.19\textwidth}
        \centering
        \includegraphics[width=\linewidth]{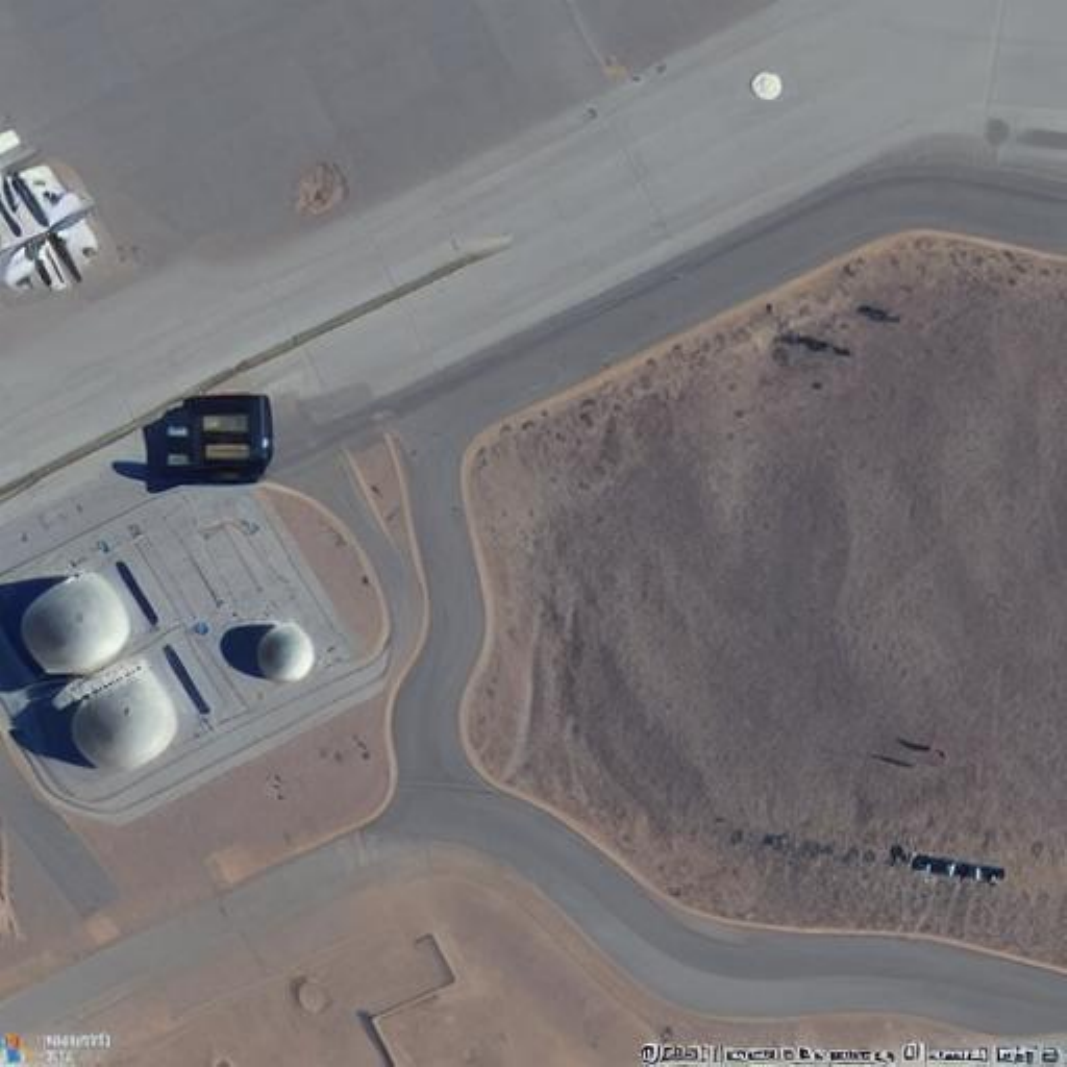}
    \end{subfigure}
    {\footnotesize\textit{"A military base in the desert, with an airport and tanks."}}
    \par\medskip
    \begin{subfigure}[t]{0.19\textwidth}
        \centering
        \includegraphics[width=\linewidth]{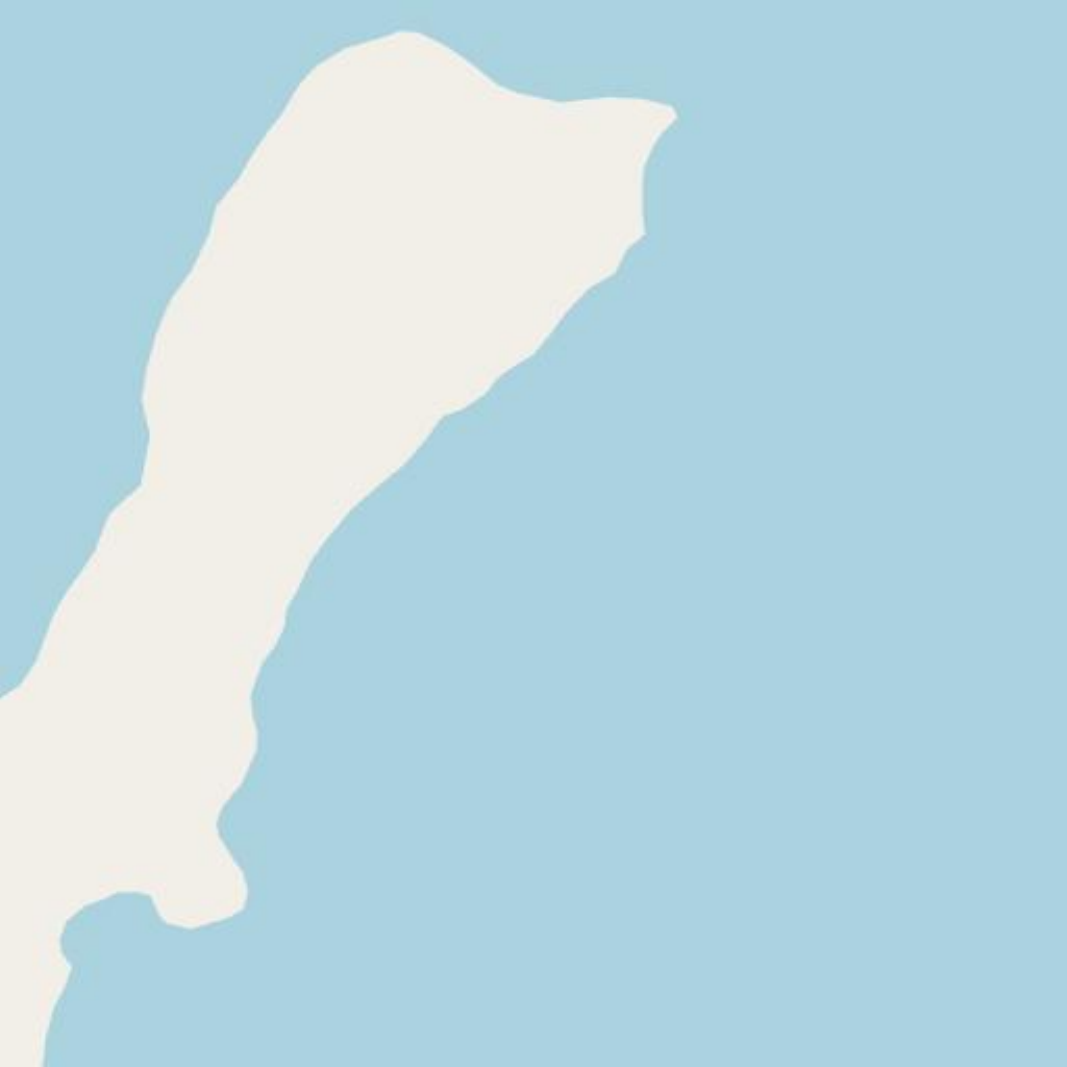}
    \end{subfigure}
    \hfill
    \begin{subfigure}[t]{0.19\textwidth}
        \centering
        \includegraphics[width=\linewidth]{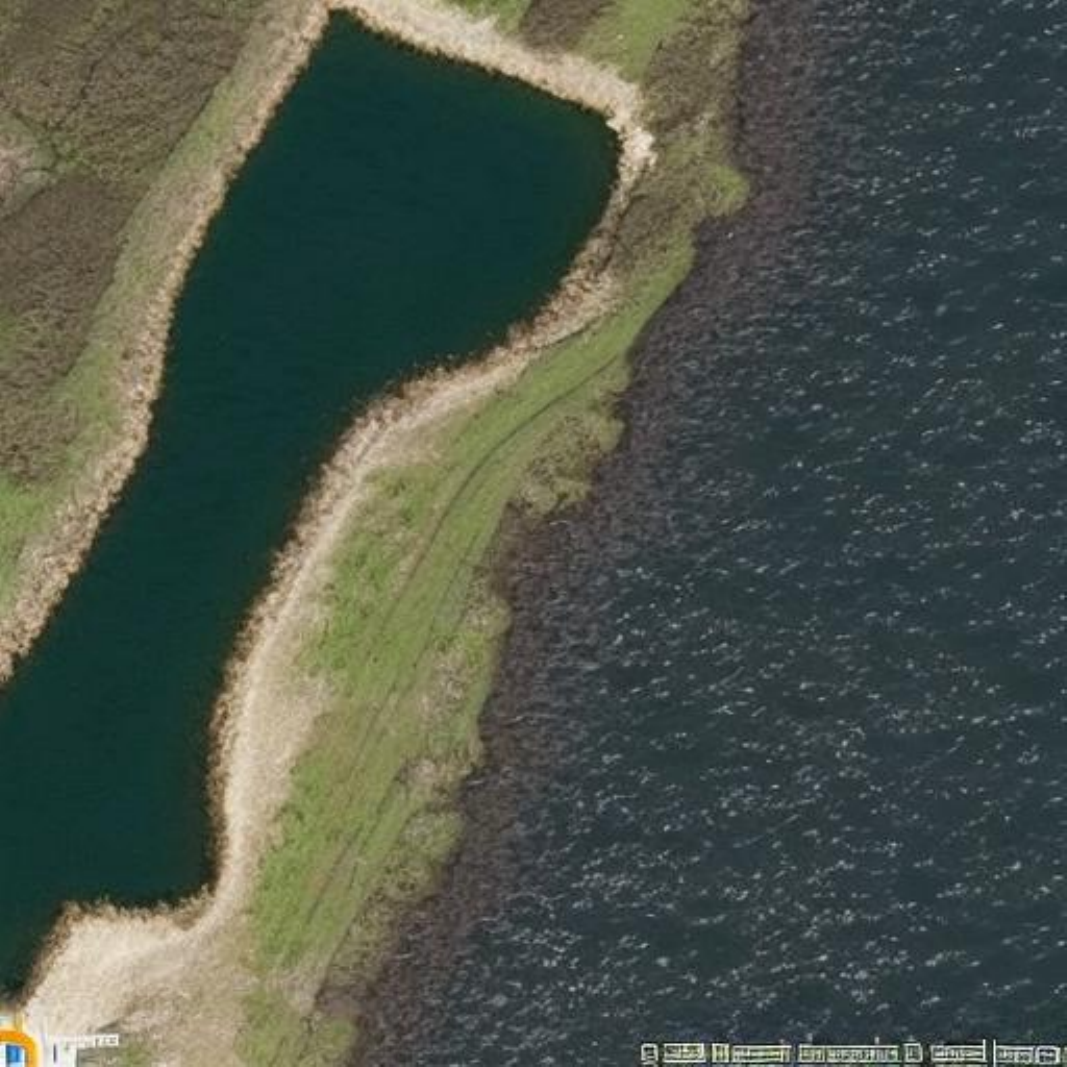}
    \end{subfigure}
    \hfill
    \begin{subfigure}[t]{0.19\textwidth}
        \centering
        \includegraphics[width=\linewidth]{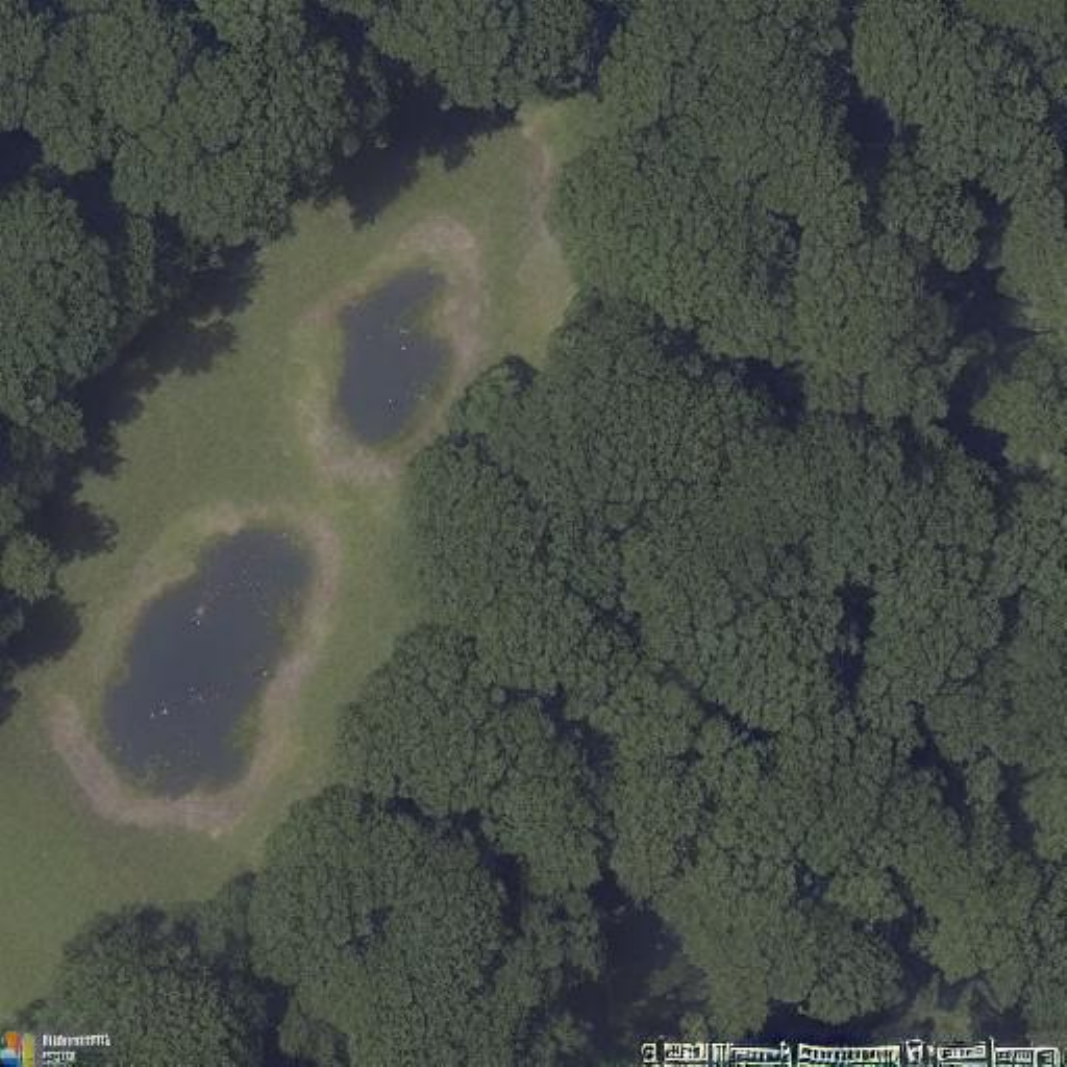}
    \end{subfigure}
    \hfill
    \begin{subfigure}[t]{0.19\textwidth}
        \centering
        \includegraphics[width=\linewidth]{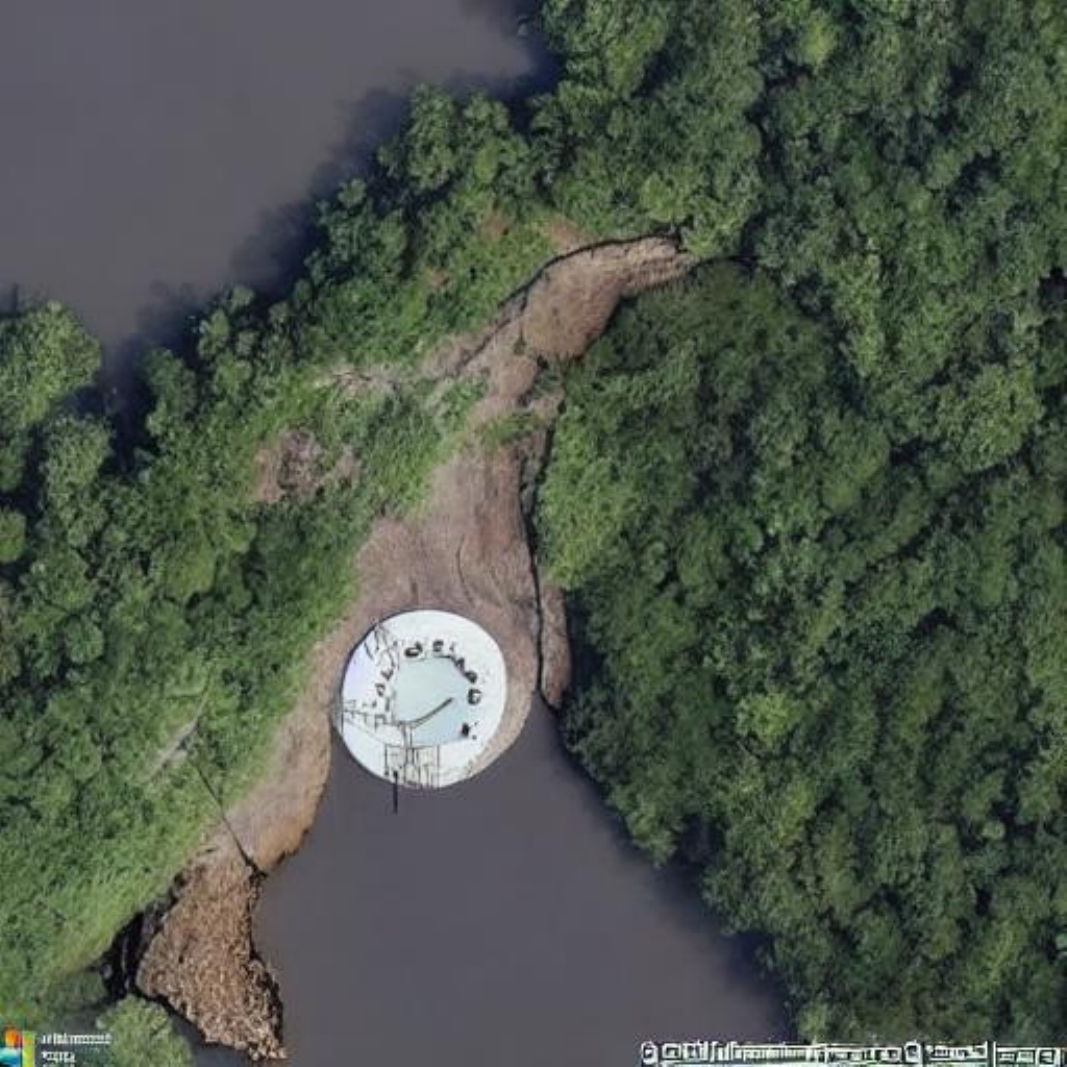}
    \end{subfigure}
    \hfill
    \begin{subfigure}[t]{0.19\textwidth}
        \centering
        \includegraphics[width=\linewidth]{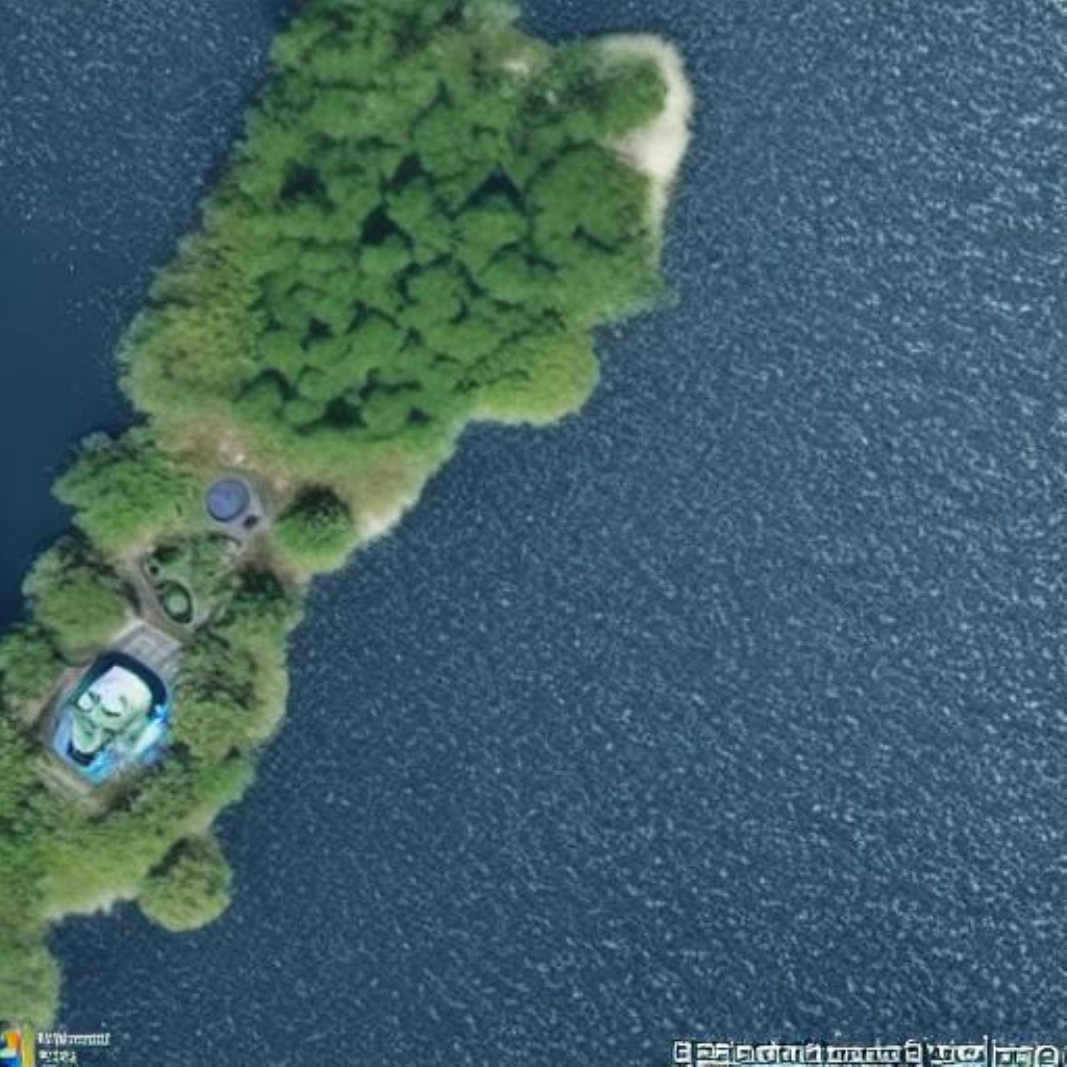}
    \end{subfigure}
    {\footnotesize\textit{"An isolated island surrounded by water, with a forest in the middle and a helipad in the upper part."}}
    \caption{Generated samples for all considered control mechanisms. OSM control maps were independently extracted to cover a wide range of scenarios.}
    \label{fig:gen-examples}
\end{figure*}

%% file: sections/conclusions.tex
\section{Conclusions}\label{sec:concl}

In this paper, we tackled the problem of adding geometric control to pre-trained text-to-image models, with the final goal of generating realistic high-resolution satellite images. Several previously-developed control techniques were tested and compared, and an additional efficient method was proposed by leveraging only skip-connection features using windowed cross-attention modules. Experiments indicate that lightweight control techniques, including our own, achieve better alignment between synthesized visual content and control signals, proving that a relatively simple and well-crafted module is sufficient to ensure good performance. In the end, we acknowledge the limitations of current metrics for quality and alignment assessment in this specialized scenario. Future research may explore this issue in greater depth, along with the development of more fine-grained control mechanisms.